\documentclass[10pt,twocolumn,letterpaper]{article}

\usepackage{iccv}
\usepackage{times}
\usepackage{epsfig}
\usepackage{graphicx}
\usepackage{amsmath}
\usepackage{amssymb}
\usepackage{multirow}
\usepackage{float}

\iccvfinalcopy % *** Uncomment this line for the final submission

\def\iccvPaperID{****} % *** Enter the ICCV Paper ID here

\ificcvfinal\fi

\begin{document}

%%%%%%%%% TITLE
\title{Estimation of Absolute Scale in Monocular SLAM Using Synthetic Data}

\author{Danila Rukhovich\\
{\tt\small drk@zurich.ibm.com}
\and
Daniel Mouritzen\\
{\tt\small dmo@zurich.ibm.com}
\and
Ralf Kaestner\\
{\tt\small alf@zurich.ibm.com}
\and
Martin Rufli\\
{\tt\small mru@zurich.ibm.com}
\and
Alexander Velizhev\\
{\tt\small ave@zurich.ibm.com}\\
IBM Research - Zurich, Switzerland
}

\maketitle
\ificcvfinal\thispagestyle{empty}\fi

%%%%%%%%% ABSTRACT
\begin{abstract}
This paper addresses the problem of scale estimation in monocular SLAM by estimating absolute distances between camera centers of consecutive image frames. These estimates would improve the overall performance of classical (not deep) SLAM systems and allow metric feature locations to be recovered from a single monocular camera. We propose several network architectures that lead to an improvement of scale estimation accuracy over the state of the art. In addition, we exploit a possibility to train the neural network only with synthetic data derived from a computer graphics simulator. Our key insight is that, using only synthetic training inputs, we can achieve similar scale estimation accuracy as that obtained from real data. This fact indicates that fully annotated simulated data is a viable alternative to existing deep-learning-based SLAM systems trained on real (unlabeled) data. Our experiments with unsupervised domain adaptation also show that the difference in visual appearance between simulated and real data does not affect scale estimation results. Our method operates with low-resolution images (0.03~MP), which makes it practical for real-time SLAM applications with a monocular camera.
\end{abstract}

%%%%%%%%% BODY TEXT
\section{Introduction} \label{Introduction}

Monocular visual SLAM allows the translation and rotation of a moving camera and a sparse map representation to be determined, but only up to scale. This paper targets the problem of absolute scale estimation using sequences of images captured by a monocular camera. We consider the most general formulation of this problem but make no assumptions about the observed scene and its objects.

Estimation of scale is important for several reasons. First, it helps recover metric feature locations, which is valuable for numerous real-world applications. Other applications that do not require absolute scale also benefit from its estimation because negative effects caused by scale drift are reduced. For example, scale drift makes it more difficult to detect loops, and selecting key frames might not take into account actual camera displacements. In addition, scale drift does not allow scale correction by optimizing just one global scale parameter.

Additional sensors might be used to overcome the scale estimation problem, including IMU, GPS, LiDAR, stereo or depth cameras. However, they raise the cost, complexity, power consumption and weight of the entire system and thus reduce the number of possible applications. Therefore, the ability to estimate scale using but a single camera would be very beneficial.

The success of deep learning has opened new options for SLAM systems. For example, single image depth prediction makes it possible to retrieve absolute depth maps from color images. This might solve the scale estimation problem as well and improves SLAM systems in general~\cite{tateno2017cvpr, yang2018_dvso}. However, this approach solves a much more complex problem, requires a lot of training data and therefore might be considered too computationally expensive for a given scale estimation problem. End-to-end deep learning approaches for SLAM were also recently introduced~\cite{DeepVO, li2019relative}. In this case, absolute scale might be estimated as well if relative camera poses with absolute scale are used for training. For the time being, complete end-to-end SLAM approaches require a modern GPU unit, which limits applications of these methods.

In this paper, we focus only on estimating the scale, which might then be integrated into classical SLAM systems. Comparison of classical and deep learning-based SLAM systems or solving the full six-degrees-of-freedom (DOF) pose estimation problem is beyond the scope of this work.

An elegant and lightweight approach for scale estimation using CNNs is described in~\cite{3dim_FrostMP17}. The absolute distance (or speed) between two consecutive images is estimated independently for each pair. This helps reduce the scale drift effect significantly. The approach is very general (no explicit assumptions are made about the observed scene or objects inside it), is applicable to monocular SLAM and can be easily integrated into existing SLAM systems. This paper considers the method established in~\cite{3dim_FrostMP17} as a baseline.

Our detailed analysis of results~\cite{3dim_FrostMP17} on the KITTI dataset~\cite{Geiger2012} highlighted several issues. First, we observed considerable systematic errors as the camera turned, see Figure~\ref{fig:TrajectoriesWithErrors}, which might be attributed to a strong domination of pure forward camera movements in the training set. Another issue is related to the sensor configuration. Specifically, a camera placed at an offset to the vehicle’s center of rotation constrains the possible combinations of rotation and translation values. For example, a front camera would never experience a pure rotation around the camera center because the global rotational center of a conventional vehicle is located on the rear wheel axis~\cite{scaramuzza2009absolute}. This means that actual camera rotation will always be combined with a certain translational movement. Adapting to new sensor configurations will require new data collection efforts. The last issue pertains to absolute scale estimation accuracy. There are often significant changes in the estimated distances between consecutive image pairs, even when the speed is relatively constant. Based on these arguments, the following goals of this paper can be formulated as follows:
\begin{enumerate}
\setlength\itemsep{0em}
\item Improve the state-of-the-art accuracy of absolute scale estimation
\item Improve estimates for camera turning movements
\item Improve robustness against variations of the sensor configuration and environmental conditions.
\end{enumerate}

Our solution to the first and the second problems is to modify the neural network architecture with respect to the baseline approach~\cite{3dim_FrostMP17} by increasing complexity of the network and adding a recurrent neural layer. These network modifications lead to significant reduction of the training loss to the very low values. This fact makes unnecessary further investigation of the network architecture, so we do not focus on this topic in the paper. The third issue is solved by using autonomous driving simulators, with which we can easily model any type of sensor configuration and generate a large amount of training data with wide variations of lighting and weather conditions. Recent works in similar domains (depth prediction~\cite{dur24333} or optical flow generation~\cite{DFIB15}) show promising results when synthetic data is used. Although our simulated training environment differs significantly from the KITTI dataset~\cite{Geiger2012} (for example, there are no parked cars), the trained network is still able to generalize to new and unseen environments. We will show that training on synthetic data yields comparable results to models trained on real data. In addition, we evaluate the influence of image photorealism for the problem at hand.

The paper is organized as follows. Section~\ref{RelatedWork} presents an overview of existing methods, and
Section~\ref{ProposedNetwork} describes proposed network architectures, datasets and domain adaptation strategies.
Section~\ref{Experiments} presents evaluation results and compares different methods.
Concluding remarks and discussion are available in Section~\ref{Conclusions}.

%-------------------------------------------------------------------------
\section{Related work} \label{RelatedWork}

This section describes previous work on the scale estimation problem for monocular SLAM. Our discussion covers the following approaches: (i) using explicit knowledge about the scene, (ii) using explicit soft assumptions about the scene, (iii) general, assumption-free approaches.

The first approach explicitly uses information about the scene, \eg~a 3D model with absolute scale. The scale of image-based reconstruction is estimated using correspondences between images and the 3D model. A known 3D model is a very strong assumption, which strongly limits the applicability of these approaches. These methods were introduced more than 30 years ago~\cite{Torlegard1988} and are mentioned primarily for completeness.

The second approach uses soft assumptions about the observed scene. This might be a known height of the camera above the ground plane~\cite{conf_ivs_GraterSL15, geiger2011stereoscan, geiger20133d, zhou2019ground} or information about absolute sizes of objects presented in a scene~\cite{conficraFrostKM16, confivsKittGL10, Scaramuzza2009, confcvprSongC14, CastleICRA2007}. In the first case, a position of the ground plane is estimated from space 3D points of the reconstructed scene, and the distance between camera center and this plane is constrained. 
In the second case, a pretrained object detector for a defined set of object classes (\eg~cars) is employed to incorporate general knowledge of object sizes into the optimization problem. 

The main disadvantage of both these approaches are the assumptions themselves. For instance, the constraint on the camera height is applicable only for cameras mounted on vehicles and assumes this height is known beforehand and is constant during image recording. Analogously, relying on having certain classes of objects in a scene fails when none of these objects are present. Moreover, observed objects might be only partially visible, which makes their size estimates inaccurate. Objects might also have intra-class variations in size (\eg~different types of cars), which additionally decreases the accuracy of the scale estimates. In order to mitigate those issues, different flexible schemes using object detection have been introduced. For example, \cite{conficraFrostKM16} introduces so-called object bundle adjustment, which optimises 3D landmark positions associated with objects of known size. The work in~\cite{conficraSucarH18} fuses single detections from a generic object detector within a Bayesian framework. Using the nonholonomic constraints of wheeled vehicles (\eg~cars, bikes or differential drive robots) was proposed in~\cite{scaramuzza2009absolute} to estimate the absolute scale from a single vehicle-mounted camera. This approach uses no assumptions about the scene, but it works only for cameras mounted on wheeled vehicles and only when the vehicle is turning. This limitation makes it difficult to recover absolute scale for long, linear trajectories.

The last type of methods is more generic and does not use explicit assumptions about the scene. The most common idea is to recover scale using absolute depth maps constructed from single images using deep-learning methods~\cite{kerl13icra, Izadi2011}. These approaches are similar to 3D reconstruction using RGB-D sensors, which directly output metric depth maps. The scale is recovered natively by integrating the absolute depth maps into the 3D reconstruction. Recent progress in depth prediction from single images~\cite{NIPS2014_5539, fu2018deep, dur24333} has made it possible to apply these methods to monocular scenes~\cite{tateno2017cvpr}.

Another approach, introduced by Frost \etal~\cite{3dim_FrostMP17}, trains the network to predict the absolute distance between camera centers from a pair of images with significant visual overlap. This approach is fairly generic because no explicit assumptions about the scene are made. Intuitively this approach learns how similar image regions are shifted between two frames. Given intrinsic camera parameters, these shifts would be proportional to the camera displacement. The distance predictions are directly included in the bundle adjustment, which improves scale accuracy significantly. The method is applied to images with a relatively low resolution of $240 \times 120$ pixels and can be executed in real time. Frost \etal~\cite{3dim_FrostMP17} also present benefits of integrating the scale estimator into a full monocular SLAM system. From our point of view this integration is straightforward and we are focusing only on improving the scale estimator itself and leaving the integration for future work.

End-to-end SLAM approaches such as~\cite{DeepVO, li2019relative} provide an alternative to the classical feature-based SLAM approach. These methods also provide full camera poses in absolute scale. We consider these approaches to be overcomplicated (in terms of the number of parameters) for dealing with the scale estimation problem alone. For example, the DeepVO method~\cite{DeepVO} introduces a fully connected layer with 122M trainable parameters and thus significantly increases computational complexity with strong implications to real-time applications.

The idea to use synthetic data for scale estimation raises the following question: How important is the visual similarity between simulated and real environments? To answer this question, one could apply domain adaptation methods, which help improve the visual realism of synthetic data and train the scale estimator with more realistic synthetic data. Numerous unsupervised image domain adaptation methods have been introduced recently (\eg~\cite{zhu2017unpaired, huang2018multimodal}) and show impressive results with regard to changing the visual similarity between different image domains (\eg~synthetic and real). The scale estimation problem takes as input a pair of images and implicitly operates on image changes so the role of photorealism itself is unclear. To evaluate this influence, we test two modern domain adaptation methods~(\cite{hoffman2017cycada, zheng2018t2net}), which are trained in an end-to-end fashion including the target problem (in our case: scale estimation). This approach allows us to optimize the visual image appearance and scale estimation within the same network. Theoretically, this allows us to apply only those image transformations that are relevant for the target problem instead of targeting visually appealing image photorealism for humans. Our implementation of both domain adaptation methods is similar to that proposed in the original papers, see~\cite{hoffman2017cycada, zheng2018t2net} for technical details.

%-------------------------------------------------------------------------
\section{Proposed method}\label{ProposedNetwork}
We will first introduce the synthetic data collection pipeline, then  propose network architectures and finally describe the domain adaptation methods we used.

\subsection{ Synthetic data collection}
\begin{figure}[H]
\begin{center}
\includegraphics[width=0.8\linewidth]{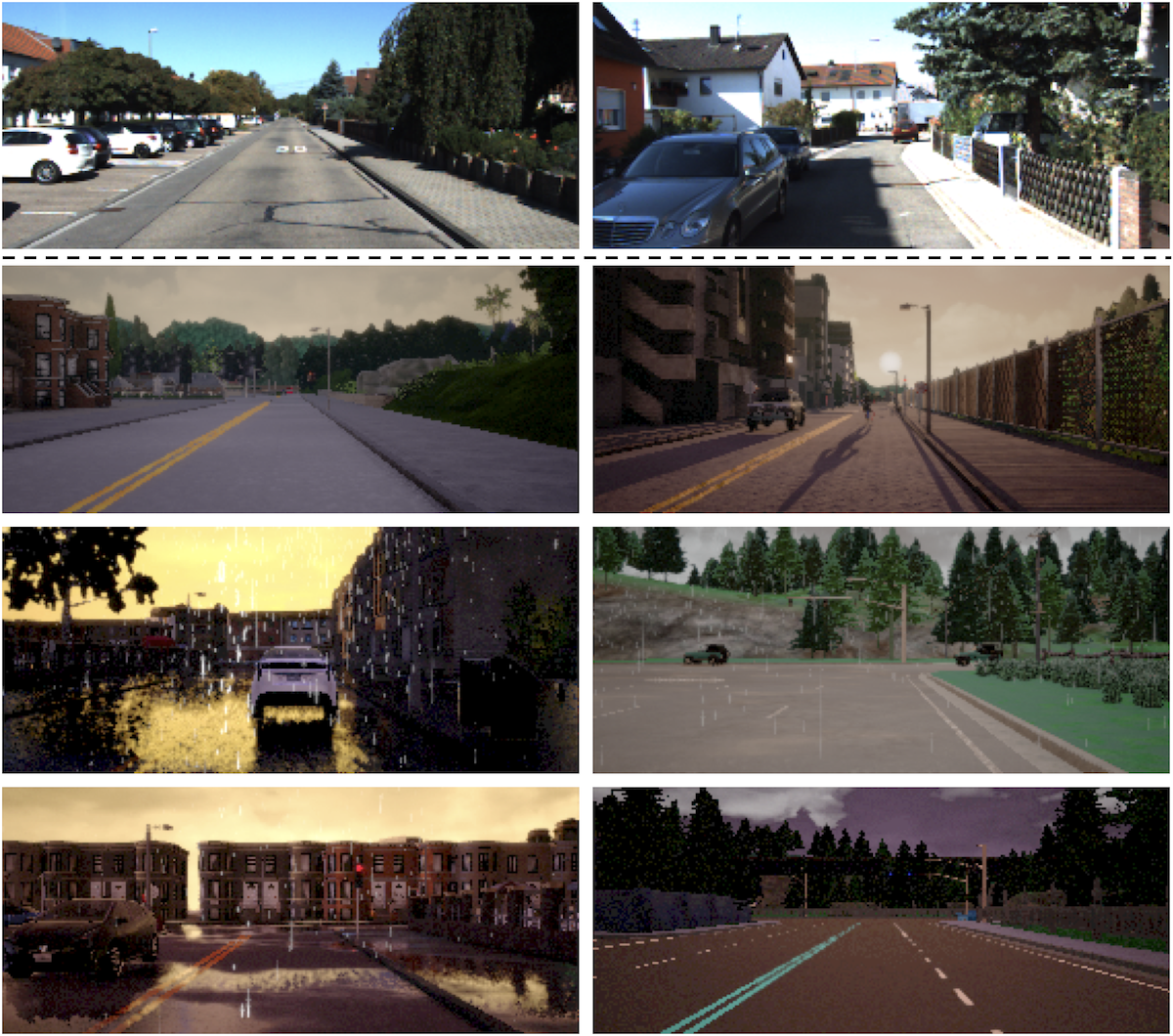}
\end{center}
 \caption{Example of real images from KITTI~\cite{Geiger2012} (top row) and CARLA simulator~\cite{Dosovitskiy17} (bottom three rows) with different daytime and weather conditions.}
\label{fig:ImageExamples}
\end{figure}

\begin{figure*}[h]
\begin{center}
 \includegraphics[width=0.8\linewidth]{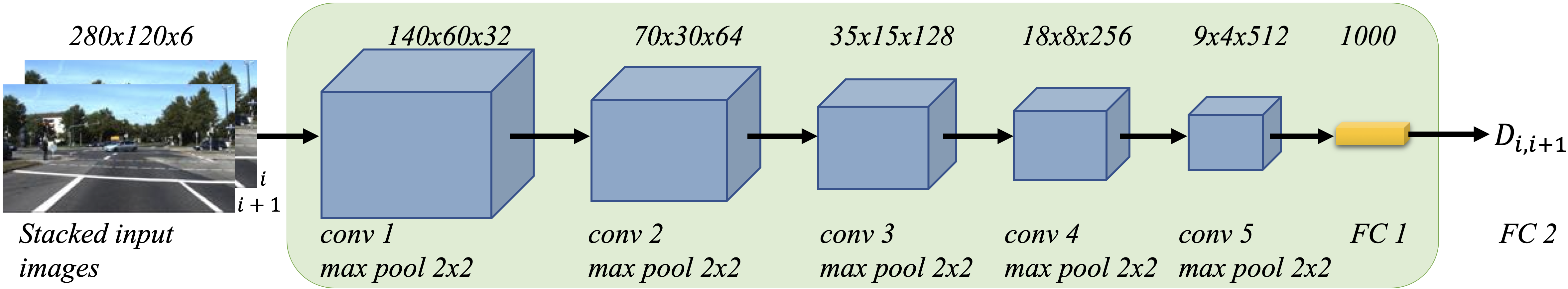}
\end{center}
 \caption{Baseline CNN architecture used in all experiments of this paper.}
\label{fig:ArchitectureCNN}
\end{figure*}

Using real data (such as the KITTI dataset~\cite{Geiger2012}) for training imposes certain limitations: the variation of weather, illumination and time of day might be limited, diversity of scene appearance (\eg urban, rural) requires significant data collection efforts. In addition, sensor configuration and vehicle movement type determine the variability of observed data in the parameter space. A change of the camera position in the vehicle could lead to the significant systematic errors because certain combinations of rotation and translation are not presented in the training set. One way to overcome these issues is the use of synthetic data collected from autonomous driving simulators like~\cite{Dosovitskiy17, airsim2017fsr, craig_quiter_2018}. A simulator allows to attach one or multiple cameras at specific locations on the vehicle, vary weather, time and lighting conditions, and define custom camera intrinsic parameters. An autopilot mode enables automatic driving in the scene combined with training data collection. In our case the collected training data includes color images and ground truth camera trajectories. The synthetic data collection pipeline and a network training procedure form one complete framework which is executed fully automatically. Basically this solves the problem of sensor configuration and, as we show later, allows to reach scale estimation accuracy comparable to accuracy produced by the model trained on real data.

\subsection{Network architectures}

To start, we selected the method from~\cite{3dim_FrostMP17} as a baseline implementation. The key idea is to concatenate a pair of RGB images into one 6-channel image, pass it through three convolutional layers with max-pooling (window size 2x2, stride 2), followed by two fully connected layers. As mentioned above, our goal is to improve the absolute distance estimation accuracy, so our first step is to increase the complexity of the network. We added two more convolutional layers, which improves the accuracy remarkably, see~Figure~\ref{fig:ArchitectureCNN} for details. In contrast to [11], we use exponential linear units (ELUs) [5] as activation functions instead of tanh(·) because it leads to a faster convergence of the training. Dropout layers are used between all convolutional and fully connected layers. A detailed configuration of the network is presented in Table~\ref{tab:ConfigurationCNN}. We use the Adam optimizer~\cite{kingma2014adam} and the mean squared error as our loss function. For simplicity, the following sections refer to this as a ``CNN'' architecture.

\begin{table}[h]
\begin{center}
\begin{tabular}{|l|c|c|c|c|}
\hline
Layer & \begin{tabular}[c]{@{}c@{}}Kernel\\Size\end{tabular} & Padding & \begin{tabular}[c]{@{}c@{}}Number of\\Filters\end{tabular}\\
\hline\hline
conv 1 & 11x11 & 5 & 32 \\ \hline
conv 2 & 9x9 & 4 & 64 \\ \hline
conv 3 & 7x7 & 3 & 128 \\ \hline
conv 4 & 5x5 & 2 & 256 \\ \hline
conv 5 & 3x3 & 1 & 512 \\ \hline
\end{tabular}
\end{center}
\caption{Configuration of CNN layers.}
\label{tab:ConfigurationCNN}
\end{table}

\begin{figure}[h]
\begin{center}
 \includegraphics[width=0.7\linewidth]{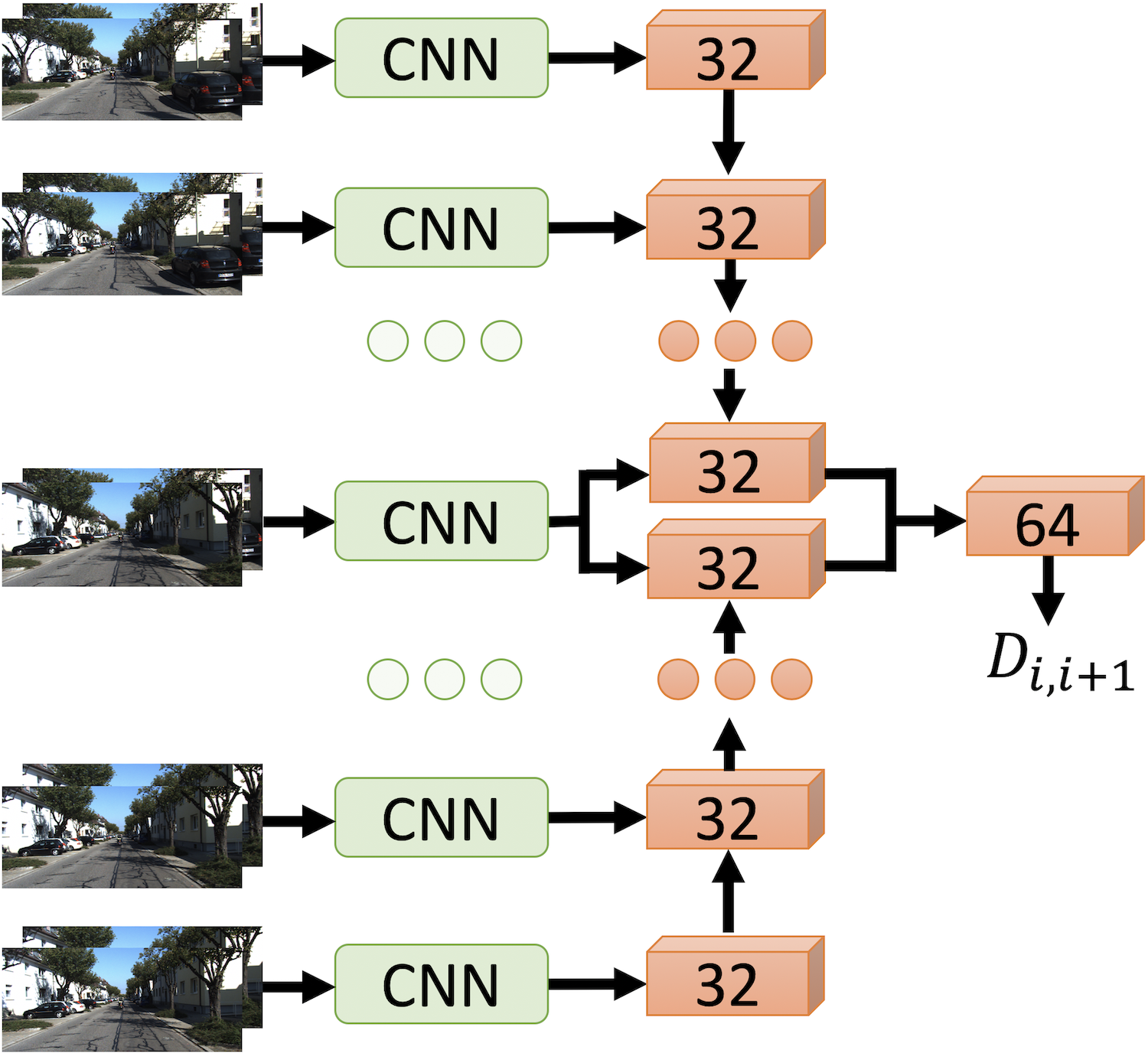}
\end{center}
 \caption{Proposed bidirectional LSTM architecture. The CNN block corresponds to the green block of Figure~\ref{fig:ArchitectureCNN}.}
\label{fig:ArchitectureLSTM}
\end{figure}

As described in Section~\ref{Introduction}, there are often significant changes in the estimated distances between consecutive image pairs. This is not surprising because distances are estimated independently for all image pairs, which means that vehicle dymanics are ignored. Instead of adding explicit constraints to possible vehicle movements, we propose that the actual dynamics be learned from data. For this purpose, we use a recurrent neural network (RNN), in particular a many-to-one LSTM~\cite{lstm}. This allows the network to learn how previously observed image pairs influence the current distance estimate. The key concept of our architecture is inspired by~\cite{lrcn2014}, which uses an LSTM for action recognition in videos. A similar approach was also used in~\cite{DeepVO} to predict full 6DOF camera pose. In contrast to the previous approaches, we also evaluate a bidirectional LSTM version, see Figure~\ref{fig:ArchitectureLSTM}). In this case, we propagate information from $N$ past and $N$ future frames. Our algorithm outputs distance estimates with a small delay, but this limitation is not critical because this step can be done in parallel with other SLAM steps such as feature detection and descriptor computation. In addition, a convolutional part of the network (the green block) is executed only once per image pair, so we need to evaluate only relatively lightweight LSTM layers.  Given the low resolution of images in our experiments ($280\times120$ pixels) this delay has a minor impact on computational performance. For comparison, we also evaluated a similar unidirectional LSTM version. We will refer to this as an ``LSTM'' architecture. 

Our changes of the baseline architecture~\cite{3dim_FrostMP17} lead to the convergence of the training loss to the values comparable to the accuracy of training data itself. So we conclude that complexity of the network is enough for the considered problem.

We use the same image normalization procedure as in~\cite{3dim_FrostMP17}, which helps make the trained model invariant to different intrinsic camera parameters. For both synthetic and real data, we use the following image augmentation steps:
\begin{itemize}
\setlength\itemsep{-0.15em}
\item Random image contrast and brightness adjustment
\item Random image horizontal flip
\item Constrained random image rotation and translation
\item Using all consecutive or non-consecutive image pairs as long as the distance between camera poses is not greater than $D\textsubscript{max}$
\item Duplication of image pairs recorded while vehicle is turning.
\end{itemize}

\subsection{Domain adaptation}
Images collected from autonomous driving simulators have quite a different appearance compared with real data, see Figure~\ref{fig:ImageExamples}. However, for the task at hand, the network takes a pair of images and, in principle, basically has to look at the geometric scene changes between the images. This raises the question of whether the difference in appearance has a significant impact on the quality of the distance estimation. To answer this question, we reimplemented and applied two state-of-the-art frameworks for unsupervised domain adaptation~\cite{hoffman2017cycada, zheng2018t2net}. Both approaches show promising results for semantic segmentation and single-image depth prediction. We trained these frameworks using a full training set of synthetic data with ground-truth camera locations and unlabeled images from the real training sequences. We tested these approaches out-of-the-box without fine-tuning them, which is beyond the scope of this paper.

%-------------------------------------------------------------------------
\section{Experiments} \label{Experiments}

\subsection{Experiment overview}
This section is organized as follows. First we describe real and synthetic datasets as well as training details for all proposed network architectures. Then, in order to evaluate the proposed methodology, we run the following series of experiments:
\begin{itemize}
\setlength\itemsep{-0.15em}
\item Comparison of absolute scale accuracy of the baseline~\cite{3dim_FrostMP17} with our proposed CNN and LSTM architectures trained on real or/and synthetic data
\item Evaluation of LSTM length
\item Evaluation of synthetic data diversity
\item Evaluation of domain adaptation.
\end{itemize}
These results are followed by a detailed analysis of errors and their distribution for our best model.

\subsection{Training details}
We observed that a large part of the image is cropped when the original intrinsic camera parameters are used~\cite{3dim_FrostMP17}. To use the full image, we take a focal length $250$~px and a principal point $(140, 60)$ for the target camera within the image normalization procedure. This change enlarges image resolution slightly from $(240, 120)$ to $(280, 120)$.

Contrast, brightness, rotation and translation are randomized with the same parameters for both images within a pair. Images are augmented and models are trained and evaluated using TensorFlow~\cite{tensorflow2015}, version~1.12. Random image rotation is applied with an angle in the range $\pm 10^{\circ}$, and random image translation is applied with a shift in the range $\pm 10\%$ of image size. We choose $D\textsubscript{max} = 1.7m$ to constrain the distance between nonconsecutive frames within a single training image pair, accounting for the maximum distance between frames in the testing set with an additional margin of $20$cm.

We follow the strategy of~\cite{3dim_FrostMP17} and use the KITTI outdoor dataset~\cite{Geiger2012} to train  and evaluate our approach using real data. Specifically, we use image sequences 01, 03, 04, 05, 06, 07, 09 and 10 for training and sequences 00, 02 and 08 for testing. Ground-truth distances between camera centers are estimated separately for the left and right-hand cameras while taking into account the different camera offsets. For the KITTI dataset~\cite{Geiger2012},  the total number of input real training image pairs before random augmentation is approximately 100~K. During evaluation, we use a total of approximately 26~K consecutive testing image pairs both the left and right-hand cameras.

\begin{table*}[h]
\begin{center}
\begin{tabular}{|c|l|c|c|c|c|c|c|c|}
\hline
\multicolumn{1}{|c|}{\multirow{2}{*}{}} & \multicolumn{1}{|c|}{\multirow{2}{*}{}} &\multicolumn{1}{|c|}{\multirow{2}{*}{}} & \multicolumn{2}{c|}{Seq \#00} & \multicolumn{2}{c|}{Seq \#02} & \multicolumn{2}{c|}{Seq \#08} \\ \cline{3-9}
\multicolumn{1}{|c|}{\#} & \multicolumn{1}{|c|}{Method} & \multicolumn{1}{|c|}{Training Data} &  \(\mu\)  &  \(\sigma\)  &  \(\mu\)  & \(\sigma\)  &  \(\mu\)  &  \(\sigma\)  \\ \hline
\hline
1 & Fixed height~\cite{geiger2011stereoscan} & KITTI only & $0.072$ & $0.252$ & $-0.012$ & $0.160$ & $0.154$ & $0.349$ \\ \hline
2 & Frost \etal \cite{3dim_FrostMP17}, CNN alone & KITTI only & $-0.014$ & $0.177$ & $-0.018$ & $0.203$ & $-0.004$ & $0.165$ \\ \hline \hline
3 & Frost \etal (our impl., $240x120$) & KITTI only & $-0.009$ & $0.177$ & $0.013$ & $0.180$ & $-0.061$ & $0.152$ \\ \hline
4 & Frost \etal (our impl., $280x120$) & KITTI only & $-0.015$ & $0.175$ & $-0.010$ & $0.178$ & $-0.057$ & $0.149$ \\ \hline \hline
5 & CNN & KITTI only & $0.009$ & $0.107$ & $0.023$ & $0.113$ & $-0.017$ & $0.092$ \\ \hline
6 & CNN & CARLA only & $-0.017$ & $0.111$ & $0.023$ & $0.105$ & $-0.029$ & $0.121$ \\ \hline
7 & CNN & CARLA and KITTI & $0.036$ & $0.079$ & $0.029$ & $0.084$ & $0.013$ & $0.081$ \\ \hline \hline
8 & CNN, LSTM (B, 19) & KITTI only & $0.019$	& $\boldsymbol{0.069}$	& $0.033$	& $\boldsymbol{0.083}$ &	$-0.017$ & $\boldsymbol{0.064}$ \\ \hline
9 & CNN, LSTM (B, 19) & CARLA only & $-0.004$ & $0.102$ & $-0.003$ & $0.132$ & $-0.049$ & $0.128$ \\ \hline
10 & CNN, LSTM (B, 19) & CARLA and KITTI & $0.039$ & $0.076$ & $0.030$ & $0.084$ & $0.002$ & $0.084$ \\ \hline
\end{tabular}
\end{center}
\caption{Comparison of absolute scale estimation results for the baseline method~\cite{3dim_FrostMP17} and proposed architectures. Means and standard deviations of the difference between ground truth and predicted values are provided separately for each testing sequence. LSTM (B, 19) stands for bidirectional LSTM with a sequence length 19. Models are trained on the KITTI dataset (KITTI only), synthetic data from the CARLA simulator~\cite{Dosovitskiy17} (CARLA only) or both datasets CARLA and KITTI. All values are in meters.}
\label{tab:ScaleEstimationResult}
\end{table*}

We chose the CARLA simulator~\cite{Dosovitskiy17} to generate synthetic training data out of convenience as it facilitates an autopilot mode. Other simulators (\cite{airsim2017fsr, craig_quiter_2018} might be applicable instead or in addition to achieve an ever larger variability of the data. The CARLA simulator allows the time of day and weather conditions to be changed---in particular, we can add puddles and vary rain intensity---and provides six different maps. These maps have different visual appearances and road networks. To collect data, we randomly initialize the vehicle's position and the weather conditions. We save all images and their respective camera poses while driving for a short distance (\eg~100~meters) in autopilot mode, then resume driving with a new vehicle position and weather settings. For performance reasons, we change maps only after 100~K image pairs have been collected. We model both the left and right RGB camera using the same extrinsic parameters, i.e., the same offsets to the vehicle’s center of rotation, as those provided for the KITTI dataset. The full synthetic training set includes $~800$~K  image pairs from all six virtual maps available in the CARLA simulator~\cite{Dosovitskiy17}.

For the CNN architecture, the learning rate of the Adam Optimizer~\cite{kingma2014adam} is set to 0.0001 and decreased by a factor of 2 after every 10~K iterations. We add a dropout layer with a probability rate of 15\%. The batch size is 75 and number of iterations is 100~K.

For the network with an LSTM part, we use slightly different parameters: initial learning rate is 0.00002 with a decay factor of 2 after every 2500 iterations. The batch size is set to 16. The augmentation scheme is as described above. The convolutional part of the network is initialized from the model trained without LSTM layers, which allows for limiting the total number of training iterations to 15~K.

\subsection{Evaluation of scale estimators}
Table~\ref{tab:ScaleEstimationResult} contains the main results of this paper, including a comparison with previous results. For each method we compute standard deviations \(\sigma\) of the difference between ground truth and predicted values. For most of the experiments means are close to zero which indicates an absence of systematic errors. We do not use more advanced metrics (like Absolute Trajectory Error (ATE)) because we estimate only distances between consecutive pairs and not full camera poses.

\begin{figure*}[h]
\begin{center}
 \includegraphics[width=0.8\linewidth]{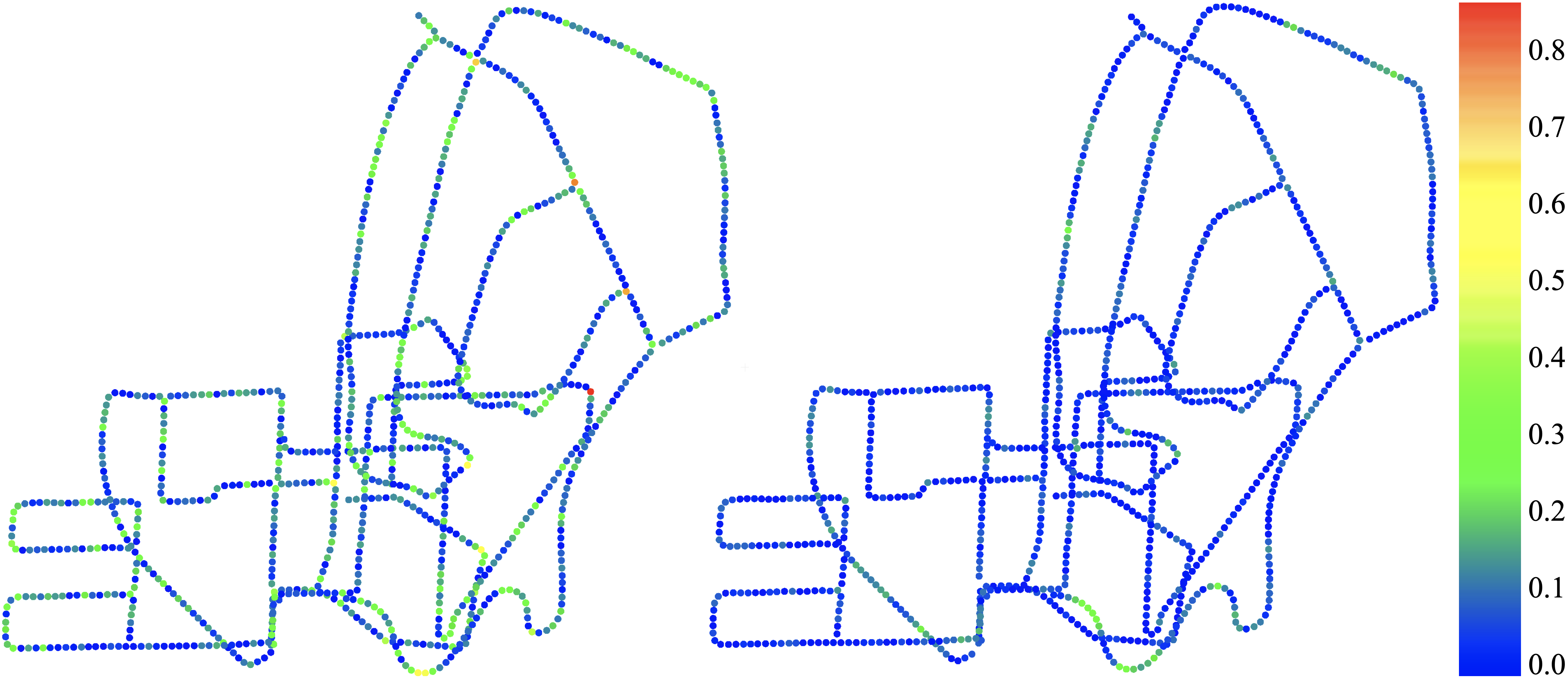}
\end{center}
 \caption{Trajectories colored by the absolute distance estimation errors (KITTI sequences 00, 02, 08). Left:~State-of-the-art results of~\cite{3dim_FrostMP17} (reimplementation). Right:~Best results of this paper. Color bar units are meters.}
\label{fig:TrajectoriesWithErrors}
\end{figure*}

The first two results come from the literature. We use slightly different settings for dropout and image augmentation than those used in~\cite{3dim_FrostMP17}, so we evaluate the influence of these changes by reimplementing that architecture. As reported within the 3rd row of Table~\ref{tab:ScaleEstimationResult}, our results are slightly better. The change of intrinsic parameters has a minor effect on the evaluation results (see rows 3 and 4), which indicates that high accuracy may be reached even with a smaller field of view.

Result 5 corresponds to our findings regarding the proposed CNN network trained using only the KITTI data, for which we observe a significant improvement of  \(\sigma\) compared to the results of~\cite{3dim_FrostMP17}. Row 6 of Table~\ref{tab:ScaleEstimationResult} presents results of the proposed CNN network trained only on synthetic data. An important conclusion comes from comparing the numbers provided in rows 5 and 6, which are quite similar. This proves the ability of the model trained on synthetic data to generalize on a completely unseen set of real images. We also trained the model using a combined dataset from KITTI and CARLA images (row 7). Clearly, these results outperform those trained using a single input modality, which means KITTI and CARLA datasets are complementary.

Our best results are obtained using a bidirectional LSTM with a length of 19, see row 8 in Table~\ref{tab:ScaleEstimationResult}. This confirms the importance of taking vehicle dynamics into account. However, in the case of CARLA-generated images, our LSTM does not improve accuracy, see rows 9--10. We interpret these results as a special property of synthetic trajectories: they are very smooth and regular, while real trajectories are more noisy and less linear. This gives rise to the assumption that our virtual vehicle dynamics do not generalize well to the real dynamics. Updating the autopilot mode of the CARLA simulator~\cite{Dosovitskiy17} might help solving this problem.

\begin{table}[h]
\begin{center}
\begin{tabular}{|c|c|c|c|c|}
\hline
Method & Training data & \(\mu\)  & \(\sigma\) \\ \hline
\hline
CNN & CARLA only & $-0.009$ & $0.109$ \\ \hline
CNN, LSTM (B, 19) & CARLA only & $0.006$ & $0.118$ \\ \hline
\end{tabular}
\end{center}
\caption{Evaluation of the models trained exclusively on CARLA data on \textbf{all available} KITTI sequences (~20K image pairs). All \(\sigma\) values are in meters.}
\label{tab:KittiEvaluation}
\end{table}

In addition we evaluated models trained exclusively on synthetic CARLA data on all KITTI sequences as none of them is used for training. Results are presented in the Table~\ref{tab:KittiEvaluation} and show that the trained models are able to generalize very well to the full corpus of KITTI data.

\subsection{Evaluation of LSTM length}
\begin{table}[h]
\begin{center}
\begin{tabular}{|c|c|c|c|}
\hline
\multicolumn{1}{|c|}{\multirow{2}{*}{}} & \multicolumn{1}{|c|}{\multirow{2}{*}{}} & \multicolumn{2}{c|}{KITTI Seq \#00, \#02, \#08} \\ \cline{3-4}
\multicolumn{1}{|c|}{\begin{tabular}[c]{@{}c@{}}LSTM\\Version\end{tabular}} & \multicolumn{1}{|c|}{\begin{tabular}[c]{@{}c@{}}LSTM\\Length\end{tabular}} & \multicolumn{1}{c|}{\(\mu\)} & \multicolumn{1}{c|}{\(\sigma\)} \\ \hline
\hline
U & 5 & $0.003$ & $0.087$ \\ \hline
U & 11 & $0.002$ & $0.085$ \\ \hline
U & 19 & $0.001$ & $0.088$ \\ \hline
\hline
B & 5 & $0.019$ & $0.084$ \\ \hline
B & 11 & $0.005$ & $0.077$ \\ \hline
B & 19 & $0.012$ & \textbf{$\boldsymbol{0.076}$} \\ \hline
\end{tabular}
\end{center}
\caption{Evaluation of different sequence lengths within the  unidirectional (U) vs bidirectional (B) LSTM arhitectures. Training and evaluation are performed on the KITTI dataset. All \(\sigma\) values are in meters.}
\label{tab:LengthLSTM}
\end{table}

Evaluation of different lengths of LSTM sequences and comparison of unidirectional and bidirectional LSTM versions are shown in Table~\ref{tab:LengthLSTM}. This experiment shows that the length of sequences plays a minor role, both for unidirectional LSTM~(U) and bidirectional LSTM~(B).

\subsection{Evaluation of synthetic data diversity}
Table \ref{tab:SyntheticSets} presents evaluation results illustrating the importance of diversity in the synthetic training data. The numbers show a clear improvement as soon as a virtual scene (or map) with appearances very different from those of the previously considered maps is added to the training set. Town 1 and 2 are quite similar to each other, so adding the data from town 2 does not yield a noticeable improvement. Town 3 contains roads with multiple lanes and thus helps to reduce \(\sigma\) values by about 1 cm. Town 6 contains two highways and improves results even further. All maps available in CARLA are not very large, and the amount of effort involved in adding more maps is small compared to the cost of launching a campaign to collect real data. This is the benefit of using synthetic data for training.

\begin{table}[h]
\begin{center}
\begin{tabular}{|l|c|c|}
\hline
\multicolumn{1}{|c|}{\multirow{2}{*}{}} & \multicolumn{2}{c|}{KITTI Seq \#00, \#02, \#08} \\ \cline{2-3}
\multicolumn{1}{|c|}{Training CARLA Maps} & \multicolumn{1}{c|}{\(\mu\)} & \multicolumn{1}{c|}{\(\sigma\)} \\ \hline
\hline
1 & $0.004$	& $0.139$ \\ \hline
1 and 2 & $-0.025$ & $0.137$ \\ \hline
1, 2 and 3 & $-0.008$ & $0.128$ \\ \hline
1, 2, 3 and 4 & $-0.012$	 & $0.129$ \\ \hline
1, 2, 3, 4 and 5 & $-0.011$ & $0.129$ \\ \hline
1, 2, 3, 4, 5 and 6 & $-0.007$ & $\boldsymbol{0.115}$ \\ \hline
\end{tabular}
\end{center}
\caption{Scale estimation results for the CNN architecture trained using synthetic data with different numbers of virtual maps. Evaluation is performed on the testing part of the KITTI dataset. All \(\sigma\) values are in meters.}
\label{tab:SyntheticSets}
\end{table}

\subsection{Results of domain adaptation}

This subsection addresses the question regarding how photorealism of synthetic data influences absolute scale estimation.
Table~\ref{tab:DomainAdaptation} contains results of distance estimations for three use cases: (i) no domain adaptation, (ii) domain adaptation using approaches T2Net~\cite{zheng2018t2net} and (iii) CyCADA~\cite{hoffman2017cycada}. Our experiments indicate almost no improvement or even minor degradation of the results if domain adaptation is applied. The difficulties may be related to the problem of changing both input images in a similar way. It seems challenging to constrain the generator of an adversarial network such that realism is equally improved on multiple images, whilst those features important for the task remain preserved. Hence, our insights are aligned with the conclusions drawn in~\cite{mayer2018makes}, i.e., that photorealism of synthetic data is less important than diversity.

\begin{table}[h]
\begin{center}
\begin{tabular}{|l|c|c|}
\hline
\multicolumn{1}{|c|}{\multirow{2}{*}{}} & \multicolumn{2}{c|}{KITTI Seq \#00, \#02, \#08}  \\ \cline{2-3}
\multicolumn{1}{|c|}{Method}  & \multicolumn{1}{c|}{\(\mu\)} & \multicolumn{1}{c|}{\(\sigma\)} \\ \hline
\hline
No domain adaptation & $-0.007$ & $\boldsymbol{0.115}$ \\ \hline
T2Net~\cite{zheng2018t2net} & $-0.011	$ & $0.117$ \\ \hline
CyCADA~\cite{hoffman2017cycada} & $0.011$ & $0.120$ \\ \hline
\end{tabular}
\end{center}
\caption{Results of domain adaptation for the CNN architecture. The training set consists of a fully annotated synthetic part and unlabeled training images from KITTI. Evaluations are performed on the testing part of the KITTI dataset. All \(\sigma\) values are in meters.}
\label{tab:DomainAdaptation}
\end{table}
	
\subsection{Error analysis}

\begin{figure}[h]
\begin{center}
\includegraphics[width=0.8\linewidth]{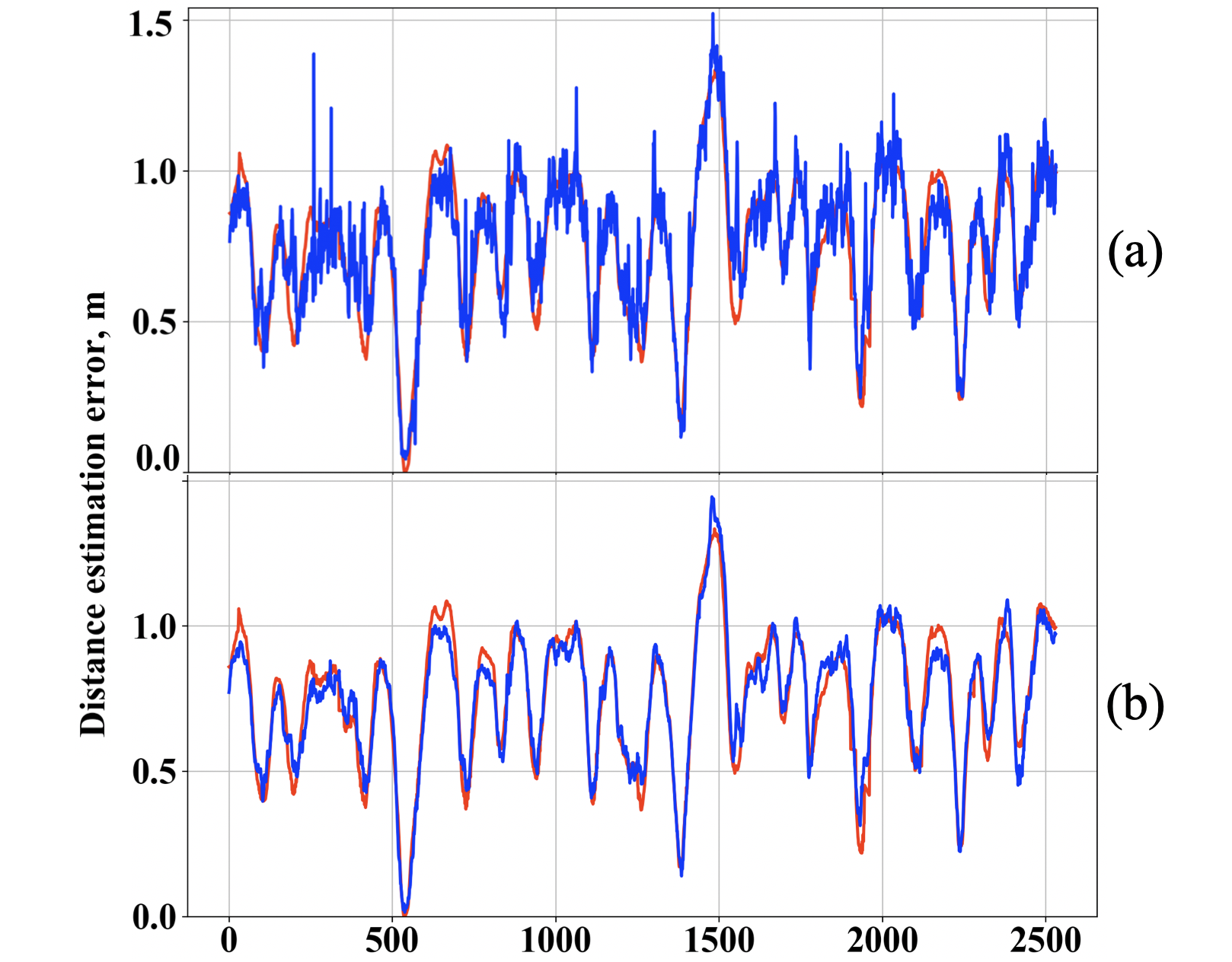}
\end{center}
 \caption{Comparison of recovered distances (blue) and ground truth (red) per frame for the part of the KITTI sequence 00 from the (a)~ CNN alone; (b)~LSTM (B), length 19. Training data: KITTI only. X-axis: frame index. }
\label{fig:LSTMInfluence}
\end{figure}

An overview of all testing trajectories for the baseline method~\cite{3dim_FrostMP17} and our best model (LSTM, length 19) is presented in Figure~\ref{fig:TrajectoriesWithErrors}. The overview gives a clear understanding of our improvements: (a)~absolute scale estimations are more accurate (b)~random noise is reduced, and (c)~large errors during vehicle turns are nearly eliminated.

Figure \ref{fig:LSTMInfluence} compares the proposed LSTM architecture with the proposed CNN architecture. Clearly, the LSTM results are smoother, which confirms the value of adding LSTM layers. However, vehicle rotations remain the most difficult challenge for the proposed method.

\begin{figure}[h]
\begin{center}
\includegraphics[width=0.8\linewidth]{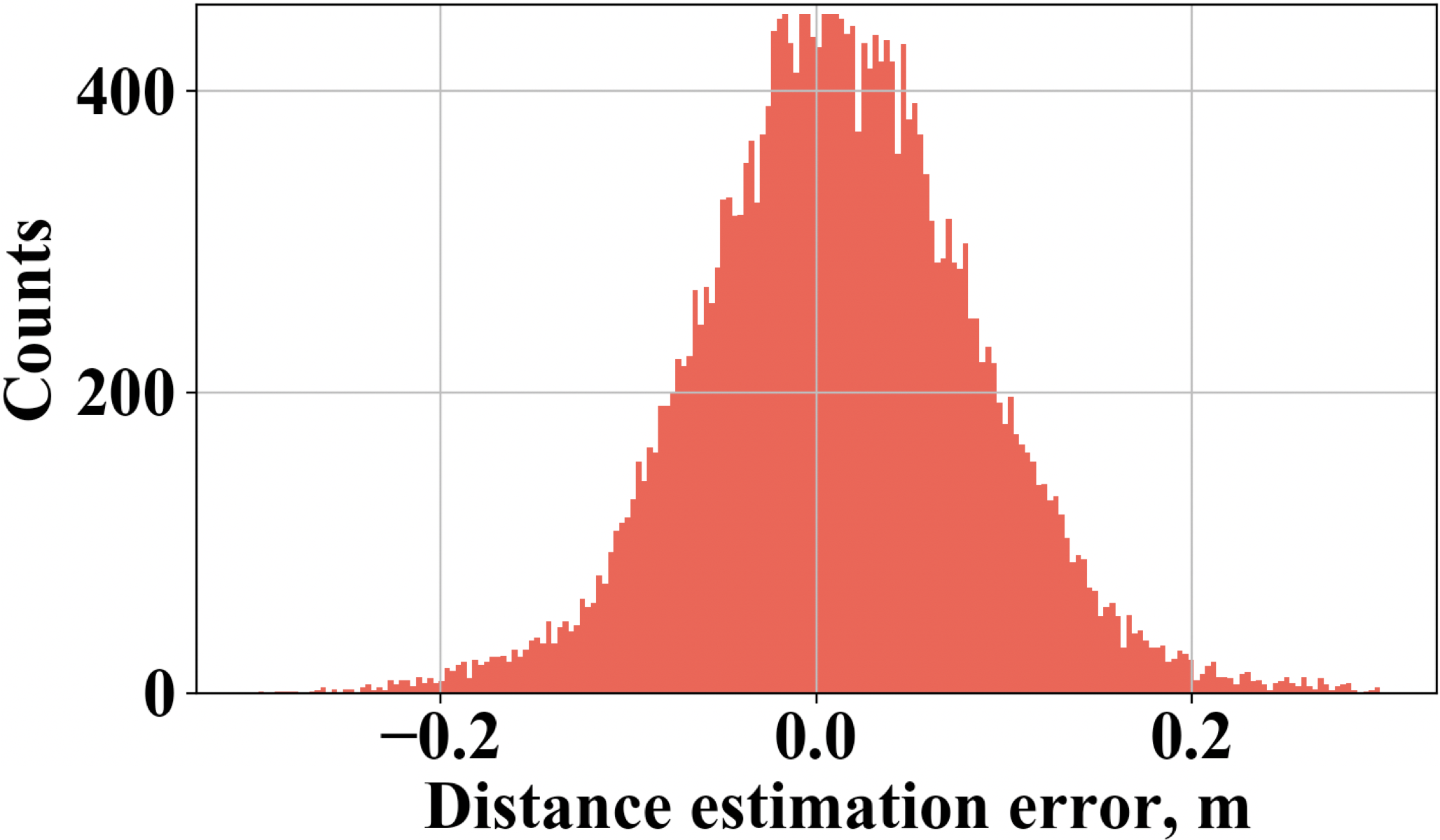}
\end{center}
 \caption{Histogram of distance estimation errors for LSTM (B, length 19) on all testing KITTI sequences. Training data: KITTI only.}
\label{fig:HistogramOfErrors}
\end{figure}

Figure~\ref{fig:HistogramOfErrors} shows the cumulated error distribution for all testing sequences. From the histogram, we observe a minor tendency of overestimating the ``true'' distances.

\begin{figure}[h]
\begin{center}
\includegraphics[width=\linewidth]{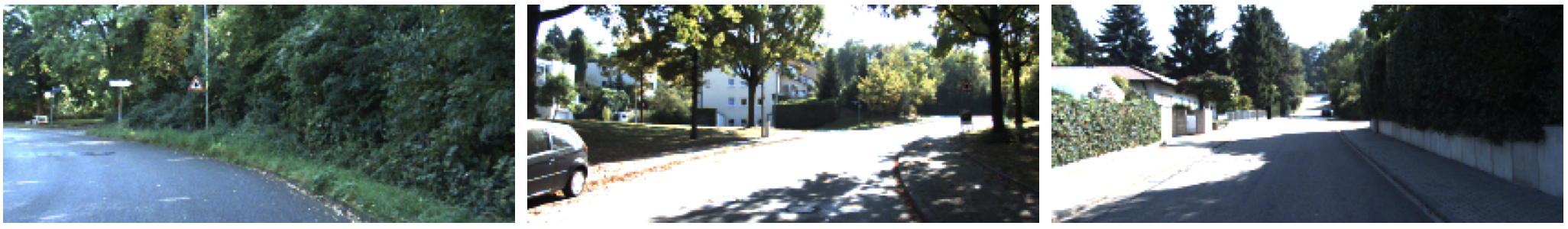}
\end{center}
 \caption{Images with largest errors for LSTM (B, length 19). Training data: KITTI only.}
\label{fig:ImagesWithErrors}
\end{figure}

Figure~\ref{fig:ImagesWithErrors} provides example images where relatively large errors are still observed (LSTM (B), length 19). The areas with large amounts of vegetation are the most difficult ones. 

%-------------------------------------------------------------------------
\section{Conclusion}\label{Conclusions}
This paper addresses the problem of scale estimation in monocular SLAM by estimating the distance between camera centers of consecutive image frames. These estimates would improve the overall performance of classical (not deep) SLAM systems and cast the entire 3D reconstruction from a monocular camera in metric values. The proposed solution estimates scale for each pair independently (or with soft-constrained LSTM network), which makes it insensitive to long-term drift effects. Our work introduced several network architectures, which lead to an improvement of scale estimation accuracy over the state of the art. With respect to the baseline method, our results show significant improvements of the estimates for camera rotations. In addition, we exploit the possibility to train the neural network only with synthetic data derived from a computer graphics simulator. Our experiments indicate that, using only synthetic training inputs, we can achieve similar scale estimation accuracy as that obtained from real data. This provides a practical solution to the sensor reconfiguration problem. Our experiments with unsupervised domain adaptation also demonstrate that differences in visual appearance (photorealism) between simulated and real data does not affect scale estimation results. The proposed methods operate with low-resolution images (0.03~MP), which makes them practical for real-time SLAM applications with a monocular camera. 

\section{Acknowledgment}\label{Acknowledgment}
This work has received funding from the European Union's Horizon 2020 research and innovation programme under grant agreement No 731993 and No 688652.

{\small
\bibliographystyle{ieee_fullname}
\bibliography{paper_for_review}
}

\end{document}